\documentclass[runningheads]{llncs}
\newcommand{\ie}{\textit{i.e.}}

\newcommand{\imdb }{\text{IMDB}}
\newcommand{\mr}{\text{MR}}

\newcommand{\lstm}{\text{LSTM}}
\newcommand{\cnn}{\text{CNN}}
\newcommand{\bert}{\text{BERT}}
\usepackage{graphicx}
\usepackage{makecell}
\usepackage{booktabs}
\usepackage{amsmath}
\usepackage{arydshln}
\usepackage[cmyk]{xcolor}
\usepackage{ulem}
\usepackage{amsfonts,amssymb}
\usepackage{bbding}
\usepackage[misc]{ifsym}
 \usepackage{color}
\usepackage{multirow}
\usepackage{algorithm}
\usepackage{algorithmic}
\begin{document}
\title{BeamAttack: Generating High-quality  Textual Adversarial Examples through Beam Search and Mixed Semantic Spaces}
\titlerunning{BeamAttack: Generating  High-quality Textual Adversarial Examples}
\author{Hai Zhu\inst{1,3,\thanks{This work was done when the author was at Ping An Technology (Shenzhen) Co., Ltd.}(}\Envelope\inst{)}\and
Qinyang Zhao\inst{2} \and
Yuren Wu\inst{3}}
\authorrunning{Hai Zhu et al.}
 %First names are abbreviated in the running head.
 %If there are more than two authors, 'et al.' is used.

\institute{University of Science and Technology of China, Hefei, China \\\email{SA21218029@mail.ustc.edu.cn} \and
Xidian University, Xi'an,  China \\\email{21151213588@stu.xidian.edu.cn} \\ \and
Ping An Technology (Shenzhen) Co., Ltd., Shenzhen, China \\
\email{wuyuren134@pingan.com.cn}}
\maketitle              % typeset the header of the contribution
\begin{abstract}
Natural language processing models based on neural networks are vulnerable to adversarial examples.   These adversarial examples are imperceptible to human readers but can mislead models to make the wrong predictions. In a black-box setting, attacker can fool the model without knowing model's parameters and architecture.  Previous works on word-level attacks  widely use single semantic space and greedy search as a search strategy. However, these methods fail to balance the attack success rate, quality of  adversarial examples and time consumption. 
In this paper, we  propose \textbf{BeamAttack}, a textual attack algorithm that  makes use of   mixed semantic spaces and improved beam search to craft high-quality adversarial examples.  Extensive experiments demonstrate that BeamAttack  can improve attack success rate while saving numerous queries and time,  e.g.,   improving at most 7\% attack success rate than greedy search when attacking the examples from MR dataset.  Compared with heuristic search, BeamAttack can save at most 85\% model queries and achieve a competitive attack success rate. The adversarial examples crafted by BeamAttack are  highly transferable and can effectively improve model's robustness during adversarial training. Code is available at \url{https://github.com/zhuhai-ustc/beamattack/tree/master}

\end{abstract}
\section{Introduction}
In recent years, neural networks have achieved great  success in the natural language processing field while being vulnerable  to adversarial examples. These adversarial examples are original inputs altered by some tiny perturbations \cite{2020-textfooler,ren-etal-2019-generating}. It is worth noting that perturbations are imperceptible to humans  but can mislead the model decision. Therefore, it is essential to explore  adversarial examples since our goal is to improve the reliability and robustness of the model, especially on some security-critical applications, such as toxic text detection and public opinion analysis \cite{yang2021besa}.  Compared to image and speech attacks \cite{szegedy2013intriguing,carlini2018audio}, it is more  challenging  in crafting textual adversarial examples due to the discrete  of natural language. In addition, there are some grammar constraints in the textual adversarial examples:(1)the crafted examples should keep the same meaning as the original texts,(2)generated examples should look natural and grammatical. However, previous works barely conform to all constraints, or satisfy the above constraints at the cost of reducing the attack success rate.  

Conventional word-level attack algorithms   can be roughly  divided into three  steps: (1) calculating  word importance score according to the changes of class label probabilities  after replacing this word,  (2) searching synonyms for each origin word, (3) selecting the substitution that reduces the class label probabilities  most  and replacing origin word  with  it until  model predicts wrong.  The problem is that previous works only use a single 
semantic space to search synonyms, which  limits the diversity of  substitutions and cut down the search space.  In addition, most prior works introduce greedy search to select the best substitution  with  the maximum change of class label probabilities \cite{2020-textfooler,li2021contextualized}. Greedy search limits the search space and sometimes leads to local optimal solution
and word over-substitution. Therefore, some works \cite{zang2020wordpso,ren-etal-2019-generating} introduce heuristic search to improve attack success rate, at the cost of   time-consuming and numerous model queries. In generally, previous works fail to balance the attack success rate, quality of  adversarial examples and time consumption.

In this paper, we propose BeamAttack, a textual attack algorithm based on mixed semantic spaces and  beam search. Specially, we search substitutions from word embedding space and BERT respectively, and filter out the bad synonyms to improve semantic similarity of adversarial examples, then  improve beam search to craft  adversarial examples, which  greatly expands the search space by controlling beam size. Therefore, it is capable of escaping from local optima within acceptable number of model queries. Furthermore, we evaluate BeamAttack by attacking various neural networks on five datasets. Experiments show that it outperforms other baselines in attack success rate and semantic similarity while saving numerous model queries.  Our main contributions are  summarized as follows:
\begin{itemize}
    \item We propose the mixed semantic spaces, making full use of  word embedding space and BERT simultaneously to  expand the  diversity of substitutions and generating high-qualify adversarial examples.
    
    \item We propose BeamAttack, a black-box attack algorithm which improves beam search to expand search space and  reduce the redundancy word substitution. 
    
    \item Experiments show that BeamAttack achieves the trade-off results compared with previous works. In addition, adversarial examples crafted by BeamAttack with high semantic similarity, low perturbation, and good transferability.
\end{itemize}
\section{Related Work}\label{sec.related_work}
 We divide the existing textual attack algorithms into char-level, word-level and sentence-level attacks based on  granularity.  Char-level attacks generate adversarial examples by inserting, swapping or removing characters(such as 'attack' $\rightarrow$ 'atttack') \cite{2019TextBugger,deepwordbug}, which can be easily  rectified by word spelling machine. Sentence-level attacks insert some perturbed sentences into the origin  paragraph to confuse models \cite{cheng2020seq2sick}. Nevertheless, these adversarial examples  contain many lexical  errors.

In order to generate high-quality adversarial examples, word-level attacks have gradually become  a prevalent approach. Word-level attacks substitute the origin words with synonyms(such as 'like' $\rightarrow$ 'love'). Traditional strategies search synonyms from word embedding space. For example, some works \cite{2020pwds,2020-textfooler,wang2022transferabletopicattack} calculate the  word saliency and greedily substitute words with synonyms derived from WordNet \cite{miller1995wordnet}, or utilizing word importance score and replace words with synonyms from counter-fitting word vectors \cite{mrksic2016counter}. Recently,  researcher  \cite{li-etal-2020-bert-attack,garg-ramakrishnan-2020-bae,li2021contextualized} 
search synonyms from  pre-trained language models (e.g. BERT, RoBERTa). The pre-trained language models are trained on massive text data, and  predict the masked words. Therefore, it has the ability to predict contextual-aware words.

Above attack algorithms adopt the greedy search, which limits the search space and leads to local optimal solution. Minor work have explored the  heuristic search, such as  genetic algorithm \cite {ren-etal-2019-generating},particle swarm optimization \cite{zang2020wordpso}.  However,  heuristic search is  very time-consuming and requires a lot of model queries. Therefore, we propose BeamAttack, searching  synonyms from word embedding space and BERT simultaneously, and fine-tuning beam search to expand search space and reduce word-over substitution.

\section{Beam Search Adversarial Attack}
\begin{figure}[t]
\centering
\includegraphics[width=9cm]{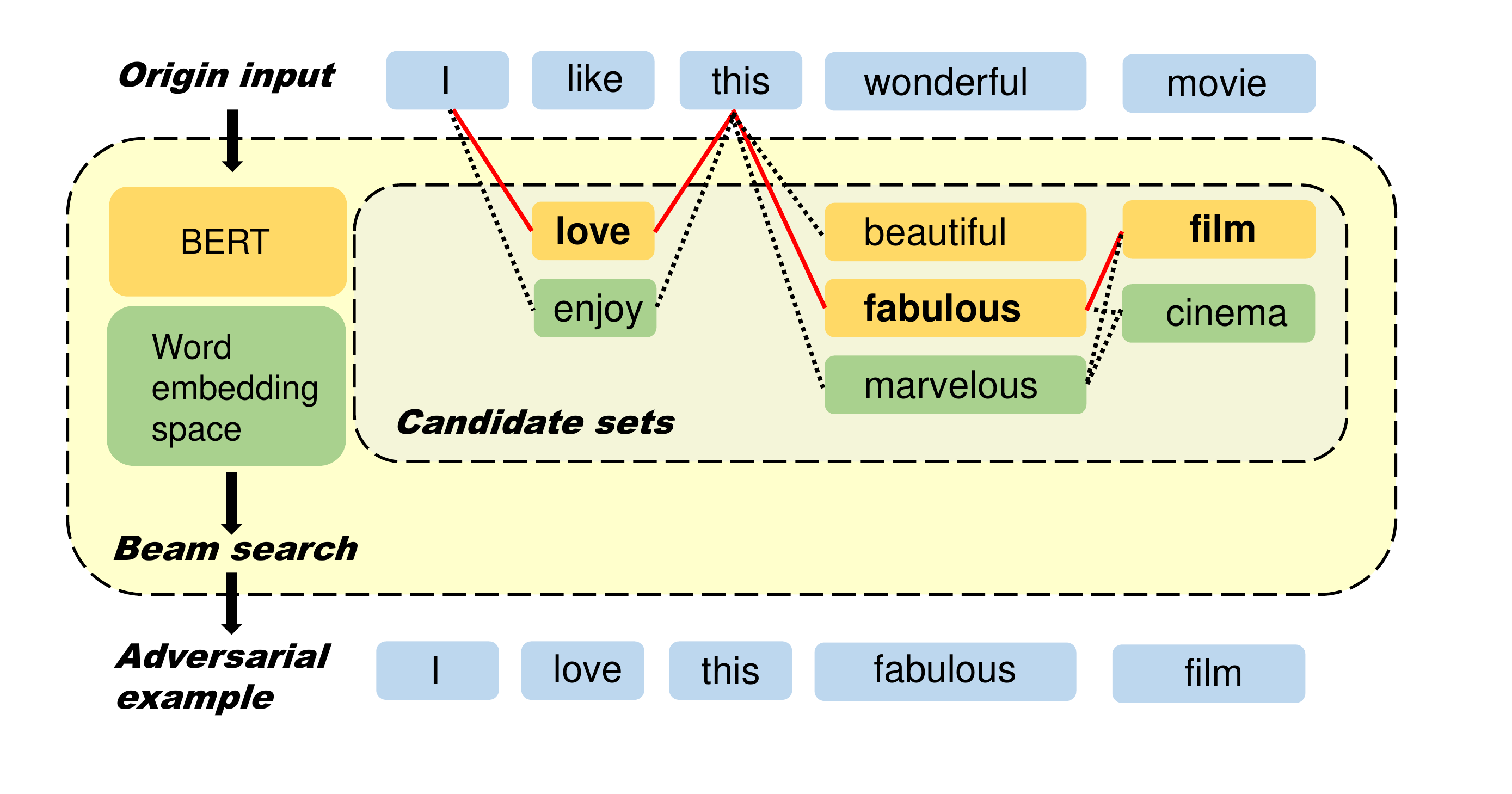}
 \vspace{-0.25in}
\caption{The overview of BeamAttack. Candidate sets are substitutions generated from BERT and word embedding space. Black lines are beam search paths, wherein red lines are the optimal search path.}
\label{fig.overriew}
\end{figure}

 BeamAttack is divided into three steps. There are word importance calculation, mixed semantic spaces and improved beam search. The overview of BeamAttack is shown in the Figure~\ref{fig.overriew}.  Before delving into details, we present the attack settings and problem formulation.
\subsection{Black-box Untargeted Attack}
The BeamAttack belongs to black-box attacks, it has nothing about model's architecture, parameters and gradients, only class label probabilities are accessible. Given a sentence of $n$ words $\mathcal{X} = [x_1,x_2,\cdots,x_n]$ and  label set $\mathcal{Y}$, a well-trained model can classify sentence correctly:
\begin{equation}
    \mathop {\text{argmax} }\limits_{y_i \in \mathcal{Y}}P(y_i| \mathcal{X}) = y_{true}
\end{equation}
 The adversarial example $\mathcal{X}' = [x^{'}_1,x^{'}_2,\cdots,x^{'}_n]$ is crafted to make model predict wrong. In addition, there are some constraints on the   word substitution rate($\mathop {\text{WSR}}$) and  semantic similarity($\mathop {\text{SIM}}$) of the adversarial example.  $\mathcal{X}'$ should be close to $\mathcal{X}$ and a human reader hardly differentiate the modifications. The mathematical expression is as follows:
\begin{equation}
\label{eq.optim}
\begin{split}
&\mathop {\text{argmax} }\limits_{y_i \in \mathcal{Y}}P(y_i| \mathcal{X'}) \neq y_{true} \\
\text{s.t.} & \mathop {\text{SIM}}(\mathcal{X'},\mathcal{X})>\mathcal{L};\mathop {\text{WSR}}(\mathcal{X'},\mathcal{X})<\sigma
\end{split}
\end{equation}
\subsection{Word Importance Calculation} Given a sentence of $n$ words $\mathcal{X} = [x_1,x_2,\cdots,x_n]$, only some important words will affect the prediction results of the model $\mathcal{F}$. In order to measure the importance of $x_i$, we follow the calculation proposed in TextFooler \cite{2020-textfooler}. We replace $x_i$ with '[oov]'\footnote{the word out-of-vocabulary.} to form $\mathcal{X}/\{x_i\} = [x_1,\cdots,x_{i-1},[oov],x_{i+1},\cdots,x_n]$, then word importance of $x_i$ is calculated as follows:

\begin{itemize}
    \item The predicted label remains the same after replace,~\ie, $\mathcal{F}(\mathcal{X})=\mathcal{F}(\mathcal{X}/\{x_i\})=y_{true}$,
\begin{equation}
\label{eq.wir1}
I(x_i)=\mathcal{F}_{y_{true}}(\mathcal{X})-\mathcal{F}_{y_{true}}(\mathcal{X}/\{x_i\})
\end{equation}
\item  The predicted label is changed after replace,~\ie, $\mathcal{F}(\mathcal{X})=y_{true} \neq y_{other} = \mathcal{F}(\mathcal{X}/\{x_i\})$,
\begin{equation}
\label{eq.wir2}
\begin{split}
I(x_i)=&\mathcal{F}_{y_{true}}(\mathcal{X})-\mathcal{F}_{y_{true}}\left(\mathcal{X}/\{x_{i}\}\right)+\\
&\mathcal{F}_{y_{other}}\left(\mathcal{X}/\{x_i\}\right)-\mathcal{F}_{y_{other}}(\mathcal{X})  
\end{split}
\end{equation}
\end{itemize}
where  $\mathcal{F}_{y}(\mathcal{X})$ represents the predicted class label  probability of $\mathcal{X}$ by $\mathcal{F}$ on label $y$. In order to improve the readability and fluency  of the adversarial examples, we will filter out stopwords by NLTK\footnote{https://www.nltk.org/} after calculating the word importance.

\subsection{Mixed Semantic Spaces} 
After ranking the words by their importance score, we need to search synonyms, which is a candidate words set $\mathcal{C}(x_i)$ for each word $x_i$. 
A proper replacement word should (i) have similar semantic meaning with original input, (ii) avoid some obvious grammar errors, (iii) and confuse model $\mathcal{F}$ to predict the wrong label. 
There are two different semantic spaces to search synonyms, word embedding spaces and pre-trained language models. 
\begin{itemize}
\item The former searches for synonyms from word embedding spaces, such as WordNet space \cite{miller1995wordnet}, HowNet space \cite{dong2010hownet} and Counter-fitting word vectors \cite{mrksic2016counter}. Word embedding spaces can quickly generate synonyms  with the same meaning as origin word.
    
\item The later searches for synonyms through pre-trained language models(such as BERT). Given a sentence of $n$ words $\mathcal{X} = [x_1,x_2,\cdots,x_n]$, we replace each word $x_i$ with '[MASK]', and get candidate words set $\mathcal{C}(x_i)$ predicted by BERT. Pre-trained language models  produce fluent and contextual-aware adversarial examples.
\end{itemize} 
We combine word embedding space and BERT to make full use of the advantage of different semantic spaces. In detail, for each word $x_i$, we respectively select top $N$ synonyms from word embedding space and BERT to form a candidate words  set  $\mathcal{C}(x_i)$. To generate high-qualify adversarial examples, we filter out the candidate words set that has different part-of-speech(POS)\footnote{
https://spacy.io/api/tagger} synonyms with $x_i$. In addition, for each $c \in \mathcal{C}(x_i)$, we substitute it for $x_i$ to generate adversarial  example $\mathcal{X'}=[x_1,\cdots,x_{i-1},c,x_{i+1},\cdots,x_n]$,  then we measure semantic similarity between  $\mathcal{X}$ and adversarial example $\mathcal{X'}$ by universal sentence encoder(USE)\footnote{https://tfhub.dev/google/ universal-sentence-encoder}, which  encodes original input  $\mathcal{X}$ and adversarial example $\mathcal{X'}$ as dense vectors and use cosine similarity as a approximation of semantic similarity. Only synonyms whose similarity is higher than threshold $L$ will be  retained in the candidate words set $\mathcal{C}(x_i)$.

\subsection{Improved Beam Search}
After filtering out the  candidate words set $\mathcal{C}(x_i)$, the construction of adversarial  examples is a combinatorial optimization problem as expected in Eq.\ref{eq.optim}. Previous works use the greedy search since it solely selects the token that maximizes the probability difference, which leads to local optima and word-over substitution.  
 
To tackle this, we improve beam search to  give consideration to both attack success rate and algorithm efficiency. Beam search has a hyper-parameter called beam size $\mathcal{K}$. 
Naive beam search only selects top $\mathcal{K}$ adversarial  examples from  the current iteration results. In the improved beam search,  we merge the output of the last iteration to the current iteration and select top  $\mathcal{K}$ adversarial examples as the input of the next iteration jointly. %The motivation for improving is considering that words are not completely independent. For example, replace words$(x_1,x_3)$ in original input may be better than $(x_1,x_2,x_3)$. Improved beam search can further reduce the redundancy word substitution. 
In detail, for each word $x_i$ in the original text, we
replace $x_i$ with the substitution from candidate words set $\mathcal{C}(x_i)$ to generate adversarial  examples  $\mathcal{X'}$ and calculate the probability differences. The top $\mathcal{K}$  adversarial  examples  $\mathcal{X'}$ with the maximum probability difference(including the last iteration of top $\mathcal{K}$ adversarial examples) are selected as the input of the next iteration until the attack succeeds or all origin words are iterated.  
It is worth noting that greedy search is a special case of  $\mathcal{K}=1$.
The details of BeamAttack are shown in algorithm \ref{algo}.
\begin{algorithm}[h] 
\caption{BeamAttack Adversarial Algorithm}
\label{algo}
\textbf{Input}:Original text $\mathcal{X}$, 
target model $\mathcal{F}$, semantic similarity threshold $\mathcal{L}=0.5$ and beam size $\mathcal{K}=10$, number of words in original text $n$\\
\textbf{Output}:Adversarial example $\mathcal{X}_{\text{adv}}$.
\begin{algorithmic}[1]
\STATE  $\mathcal{X}_{\text{adv}} \leftarrow \mathcal{X}$ \\
\STATE  $set({\mathcal{X}_\text{adv}}) \leftarrow \mathcal{X}_{\text{adv}}$ \\
\FOR{each word $x_i$ in $\mathcal{X}$}
\STATE Compute the importance score $I(x_i)$ via Eq.\ref{eq.wir1} and \ref{eq.wir2}.
\ENDFOR
\STATE Sort the words with importance score $I(x_i)$\;
\FOR{$i=1$ to $n$}

\STATE Replace the $x_i$ with [MASK]\;
\STATE Generate the candidate set $\mathcal{C}(x_i)$ from BERT and Word Embedding Space\;
\STATE $\mathcal{C}(x_i)$ $\leftarrow$ POSFilter($\mathcal{C}(x_i)$) $\cap$   USEFilter($\mathcal{C}(x_i)$)
\ENDFOR
\FOR{$\mathcal{X}_{\text{adv}}$ in $set(\mathcal{X}_{\text{adv}})$}
\FOR {$c_k$ in $\mathcal{C}(x_i)$}
\STATE $\mathcal{X'}_{\text{adv}}$ $\leftarrow$ Replace $x_i$ with $c_k$ in  $\mathcal{X}_{\text{adv}}$\;
\STATE  Add $\mathcal{X'}_{\text{adv}}$ to the $set(\mathcal{X}_{\text{adv}})$\;
\ENDFOR

\FOR{ $\mathcal{X'}_{\text{adv}}$ in $set(\mathcal{X}_{\text{adv}})$}

\IF { $\mathcal{F}(\mathcal{X'}_{\text{adv}}) \neq y_{true}$ }
\STATE \textbf{return} $\mathcal{X'}_{\text{adv}}$ with highest semantic similarity\;
\ENDIF
\ENDFOR
\STATE $set(\mathcal{X}_{\text{adv}})$ $\leftarrow$ Select top $\mathcal{K}$  adversarial examples in $set(\mathcal{X}_{\text{adv}})$\;
\STATE $i \leftarrow i+1$
\IF{$i>n$}
\STATE \textbf{break}
\ENDIF
\ENDFOR
\STATE \textbf{return} adversarial examples $\mathcal{X}_{\text{adv}}$
\end{algorithmic}
\end{algorithm}
\section{Experiments}\label{sec.experiments}
\noindent\textbf{Tasks, Datasets and Models.} To evaluate the effectiveness of BeamAttack, we conduct experiments on  two NLP tasks, including text  classification and text inference. In particular, the experiments cover various datasets, such as \mr~\cite{mr_data},\imdb~\cite{imdbdata},SST-2~\cite{socher2013recursivesst}, SNLI~\cite{bowman2015large} and MultiNLI~\cite{williams2018broad}.   We train three neural networks as target models including \cnn~\cite{kim-2014-convolutional}, \lstm~\cite{6795963} and  \bert~\cite{bert2019}. Model parameters are consistent with TextFooler's\cite{2020-textfooler} setting.

%\begin{table}[t]
%\centering
%\renewcommand\arraystretch{1.1}
%\tabcolsep=0.3cm
%\footnotesize
%\small
%\caption{Overview of datasets and NLP tasks.}
%\label{tab.dataset}
%\begin{tabular}{p{70pt}<{\centering}|p{40pt}<{\centering}p{40pt}<{\centering}p{40pt}<{\centering}p{50pt}<{\centering}}\hline
%Task                            & Dataset   & Train & Test & Length \\ \hline
%\multirow{3}{*}{Classification} & MR        & 9K    & 1K   & 18.5  \\
%                                & IMDB      & 25K   & 25K  & 235.7 \\
%                                & SST-2     & 70K   & 2K   & 8.6   \\\hline
%\multirow{2}{*}{Inference}     & SNLI      & 570K  & 3K   & 20.2  \\
%                                & MNLI  & 433K  & 10K  & 11.2  \\\hline
%\end{tabular}
%\end{table}

\noindent\textbf{Baselines.} To  quantitatively evaluate BeamAttack, we compare it with other black-box attack algorithms, including  TextFooler(TF)~\cite{2020-textfooler}, PWWS~\cite{ren-etal-2019-generating},  BAE~\cite{garg-ramakrishnan-2020-bae}, Bert-Attack(BEAT)~\cite{li-etal-2020-bert-attack} and PSO~\cite{zang2020wordpso}, wherein TF,PWWS and PSO search synonyms from word embedding spaces, BAE and BEAT search synonyms from BERT. In addition, PSO belongs to 
heuristics search and other belong to greedy search. These baselines are implemented on the TextAttack framework~\cite{morris2020textattack}.

\noindent\textbf{Automatic Evaluation Metrics.}
We evaluate the attack performance by following  metrics.  
Attack Success Rate(ASR) is  defined as the proportion
of successful adversarial examples to the total number of examples.
Word Substitution Rate(WSR) is defined as the proportion of number of replacement words to number of origin words.
Semantic Similarity(SIM) is  measured by Universal Sentence Encoder(USE).
Query Num(Query) is the number of model queries during adversarial attack. 
The ASR evaluates how successful the attack is. The WSR and semantic similarity together evaluate how semantically similar the original texts and adversarial examples are. Query num can reveal the efficiency of the attack algorithm.

\noindent\textbf{Implementation Details.} 
In our experiments, we carry out all experiments on  NVIDIA Tesla P100 16G GPU. We set the beam size $\mathcal{K}=10$, number of each candidate set $N=50$, semantic similarity threshold $\mathcal{L}=0.5$, we take the average value of 1000 examples as the final experimental result.

\subsection{Experimental Results}
The experiment results  are listed in Table~\ref{tab.asr}. It is worth noting that BeamAttack achieves higher ASR  than baselines on almost all scenarios.  BeamAttack also reduces the WSR on some datasets(MR,IMDB and SST-2). We attribute this superiority to the fine-tuned beam search, as this is the major improvement of our algorithm compared with greedy search.  BeamAttack has chance to  jump out of the local optimal solution and find out the adversarial examples with lower perturbation by expanding the search space. In terms of model robustness, BERT has better robustness than traditional classifiers(CNN and LSTM), since the attack success rate of  attacking BERT is lower than other models.
\begin{table*}[t]
\renewcommand\arraystretch{1.7}
\footnotesize
\centering
\large
\caption{The attack success rate and word substitution rate of different attack algorithms on five datasets. The “Origin ACC(\%)” denotes the target model’s test accuracy on the original inputs.}
\label{tab.asr}
\resizebox{\textwidth}{!}{
\begin{tabular}{lcc|cccccc|cccccc}
\toprule[2pt]
\multirow{2}{*}{Datasets} & \multirow{2}{*}{\makecell[c]{Target\\Models}} & \multirow{2}{*}{\makecell[c]{Origin\\ACC}} & \multicolumn{6}{c}{Attack Success Rate(ASR(\%))}                                                          & \multicolumn{6}{c}{Word Substitution Rate(WSR(\%))}                                \\ \cline{4-15} 
                      &                        &                      & TF           & PWWS         & BAE          & BEAT  & PSO          & BeamAttack       & TF    & PWWS  & BAE   & BEAT  & PSO   & BeamAttack     \\ \hline
\multirow{3}{*}{MR}   & CNN                    & 80.4                  & 98.81        & 98.61         & 98.00           & 83.31  & 96.23         & \textbf{99.87}   & 17.05 & 13.22 & 12.98 & 15.06 & 11.53 & \textbf{8.29} \\
                      & LSTM                   & 80.7                 & 98.92        & 97.92         & 98.21         & 84.12  & 95.32         & \textbf{99.90}   & 15.61 & 13.07 & 11.71 & 13.59 & 10.91 & \textbf{8.60} \\
                      & BERT                   & 90.4                 & 90.54         & 81.53         & 90.61         & 88.36  & 92.47         & \textbf{97.88}  & 20.91 & 14.67 & 14.44 & 15.32 & 11.93 & \textbf{9.70} \\ \hline
\multirow{3}{*}{IMDB} & CNN                    & 89.2                 & \textbf{100} & \textbf{100} & \textbf{100} & 99.82  & \textbf{100} & \textbf{100}     & 2.51   & 2.23   & 2.01     & 3.32  &    2.43   & \textbf{2.11} \\
                      & LSTM                   & 89.8                 & 99.76         & 99.47         & \textbf{100} & 99.83  & \textbf{100} & \textbf{100}     & 3.12   & 3.11   & \textbf{2.25}   & 3.45  &  2.46     & 2.43 \\
                      & BERT                   & 90.9                 & 88.83         & 86.55         & 83.96         & 88.68  & 89.93            & \textbf{91.6}             & 3.81   & 5.02     & 7.69   & 5.66  &    4.32   & \textbf{1.65} \\\hline
\multirow{2}{*}{SST-2}  & CNN                    & 82.5                & 92.37        & 98.23        & 95.45        & 86.44 & 96.69        & \textbf{99.88} &  17.09     &13.10       & 12.53      &  15.40 & 11.47 & \textbf{8.46} \\
                      & LSTM                   & 84.6                & 93.21        & 98.48        & 96.23         & 86.43 & 96.42        & \textbf{100}       &17.55       &13.53   &12.83 &15.31 &11.45 & \textbf{8.76}\\ \hline
SNLI                  & BERT                   & 89.1                & 96.00           & 98.42        & 98.84        & 98.64 & 92.51        & \textbf{99.80}   & 17.26  & 13.72  & \textbf{6.91}   & 7.80   & 8.19      & 13.81          \\
MNLI               & BERT                   & 85.1                & 90.44         & 94.33         & 99.23         & 92.00    & 83.43         & \textbf{99.50}   & 13.93  & 10.12  & \textbf{5.45}   &5.64       & 6.65  & 10.81         \\\bottomrule[2pt]
\end{tabular}}
\end{table*}

\begin{figure}[t]
\centering
\includegraphics[width=9cm]{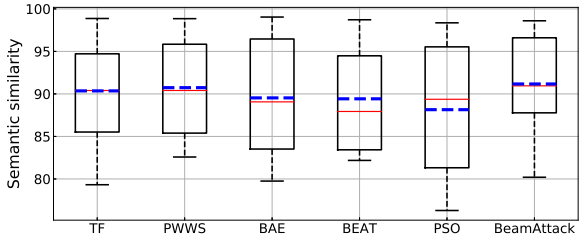}
 \vspace{-0.2in}
\caption{The semantic similarity between origin inputs and adversarial examples.}
\label{fig.semantic}
\end{figure}

\noindent\textbf{Semantic Similarity.} Except ASR and WSR, fluent and contextual-aware adversarial examples are also essential. Figure~\ref{fig.semantic} plots the semantic similarity of adversarial examples generated by different attack algorithms. Clearly, BeamAttack achieves the highest semantic similarity than other attack algorithms.

\noindent\textbf{Qualitative Examples.} To more intuitively contrast the fluency of adversarial examples, we list some adversarial examples generated by different attack algorithms in Table~\ref{tab.examples}. Compared with other methods, Beamattack not only ensures the semantic similarity between replacement words and original words but also successfully misleads the model with the minimum perturbation.

%To more intuitively contrast the fluency of adversarial examples. We list some  adversarial examples generated by different attack algorithms in Table~\ref{tab.examples}. Compared with other methods, Beamattack not only ensures the semantic similarity between  replacement words and  original words, but also successfully misleads the model with  the minimum perturbation.  Firstly,  BeamAttack makes full use of  BERT and embedding space,  providing sufficient substitutions with high semantic similarity to the original words. Secondly, BeamAttack fine-tunes beam search to expand search spaces and  reduce redundancy word substitution.

\begin{table}[t]
    \caption{The adversarial example crafted by different attack algorithms on MR(BERT) dataset.  Replacement words are represented in  \textcolor{red}{red}. }
    \small
    \centering
    \renewcommand\arraystretch{1.1}
    \label{tab.examples}
    \begin{tabular}{p{80pt}<{\centering}p{250pt}<{\centering}} \toprule[1.5pt]
      \textbf{\multirow{2}{*}{\makecell[c]{Origin Text\\ (Positive)}}} & The experience of the roles in the play makes us generates an enormous feeling of empathy for its characters.\\\hdashline
    
      \textbf{\multirow{2}{*}{\makecell[c]{BAE\\ (Negative)}}} & The experience of the roles in the play makes us generates an \textbf{\textcolor{red}{excessive}}  \textbf{\textcolor{red}{need}} of empathy for its characters.\\ \hdashline

     \textbf{\multirow{2}{*}{\makecell[c]{PWWS\\ (Negative)}}} & The experience of the roles in the play makes us  \textbf{\textcolor{red}{render}}  an enormous  \textbf{\textcolor{red}{smell}} of empathy for its  \textbf{\textcolor{red}{eccentric}}. \\ \hdashline

      \textbf{\multirow{2}{*}{\makecell[c]{TextFooler\\ (Negative)}}} & The experience of the roles in the play makes us \textbf{\textcolor{red}{leeds}} an enormous \textbf{\textcolor{red}{foreboding}} of empathy for its  \textbf{\textcolor{red}{specs}}. \\ \hdashline
      
     \textbf{\multirow{2}{*}{\makecell[c]{BeamAttack\\ (Negative)}}} & The experience of the roles in the play makes us generates an enormous feeling of \textbf{\textcolor{red}{pity}} for its characters.\\\bottomrule[1.5pt]
    
    \end{tabular}
    \end{table}

\noindent\textbf{Model Query.} The number of model queries measures the effectiveness of attack algorithm. Table~\ref{tab.query} lists the model queries of various attack algorithms. Results show that although our BeamAttack 
needs  more model queries than greedy search(such as TF), compared with the PSO attack algorithm, which adopts heuristic search,  our algorithm  obtains competitive results with extremely few model queries. 
\begin{table}[t]
\footnotesize
\centering
\small
\caption{The average model queries of different attack algorithms on five datasets. Beam size $\mathcal{K}=10$}
\label{tab.query}
%\resizebox{\linewidth}{!}{
\begin{tabular}{p{60pt}<{\centering}p{48pt}<{\centering}p{48pt}<{\centering}p{48pt}<{\centering}p{48pt}<{\centering}p{48pt}<{\centering}}\toprule[1.5pt]
 & MR     & IMDB    & SST-2  & SNLI  & MNLI   \\\hline
TF               & 113.8  & 536.7   & 146.2  & 54.1  & 68.9   \\
PWWS             & 285.4  & 3286.5  & 5054.3 & 137.7 & 157.4  \\
BAE              & 104.2  & 567.1   & 171.0  & 75.5  & 75.1   \\
BEAT             & 207.9  & 585.0   & 245.6  & 93.6  & 119.2  \\
PSO              & 5124.5 & 15564.3 & 3522.8 & 416.6 & 1124.8 \\
BeamAttack       & 650.3  & 2135.8  & 584.3  & 126.0 & 174.0  \\\bottomrule[1.5pt]
\end{tabular}
\end{table}

\subsection{Ablation Study}
\noindent\textbf{The effect of beam size $\mathcal{K}$.}
To validate the effectiveness of beam size $\mathcal{K}$, we use BERT as the target model and test on MR dataset with different beam size $\mathcal{K}$. When $\mathcal{K}=1$, beam search is equal to greedy search. As shown in Table~\ref{tab.beamsize}, the attack success rate increases  gradually with the grow of beam size $\mathcal{K}$. %We attribute this superiority to the beam search, which  is the one improvement of our BeamAttack compared with other attack algorithms(TF and BAE).

\begin{table}[t]
\caption{The effect of beam size $\mathcal{K}$ on MR(BERT) dataset.}
\footnotesize
\small
\centering
\renewcommand\arraystretch{1}
\label{tab.beamsize}
\begin{tabular}{p{60pt}<{\centering}p{40pt}<{\centering}p{40pt}<{\centering}p{60pt}<{\centering}p{60pt}<{\centering}} \toprule[1.5pt]
   Beam Size  & ASR(\%)   & WSR(\%)  &  Similarity(\%)  & Query \\ \hline
$\mathcal{K}=1$  & 89.0  & 15.5 & 81.3 & \textbf{101.3} \\
$\mathcal{K}=2$  & 90.6 & 15.3  & 82.0 & 150.4   \\
$\mathcal{K}=5$  & 91.6 & \textbf{15.1} & 82.8 & 312.6   \\
$\mathcal{K}=7$ & 92.3 & \textbf{15.1}  & 83.0 & 411.1   \\
$\mathcal{K}=10$ & \textbf{92.6} & \textbf{15.1} & \textbf{83.1} & 516.2   \\\bottomrule[1.5pt]
\end{tabular}
\end{table}

\noindent\textbf{The effect of mixed semantic spaces.}
Another major improvement of our BeamAttack is  that substitutions are selected from mixed semantic spaces.  As shown in the Table~\ref{tab.transs}, we study the impact of different  semantic spaces on different metrics. Compared with single word embedding space or BERT,  using both word embedding space and BERT to generate adversarial examples can obtain higher attack success rate,  semantic similarity and lower word substitution rate.
%\vspace{-3mm}
\begin{table}[t]
    \caption{The effect of different semantic spaces on MR(BERT) dataset.}
    \centering
    \label{tab.transs}
    \begin{tabular}{p{100pt}<{\centering}p{40pt}<{\centering}p{40pt}<{\centering}p{60pt}<{\centering}p{60pt}<{\centering}} \toprule[1.5pt]
    semantic space     & ASR(\%)     & WSR(\%) & Similarity(\%)   & Query \\ \hline
    Embedding      & 89.0  & 15.3 & 81.2 & 101.3 \\
    BERT    & 93.6  & 13.1 & 82.6 & \textbf{101.1} \\
    Embedding+BERT & \textbf{95.3} & \textbf{11.7} & \textbf{84.9} & 140.3 \\\bottomrule[1.5pt]
    \end{tabular}
    \end{table}

 %\vspace{-3mm}

 \subsection{Transferability}
\begin{figure}[h]
 \centering
 \includegraphics[width=8cm]{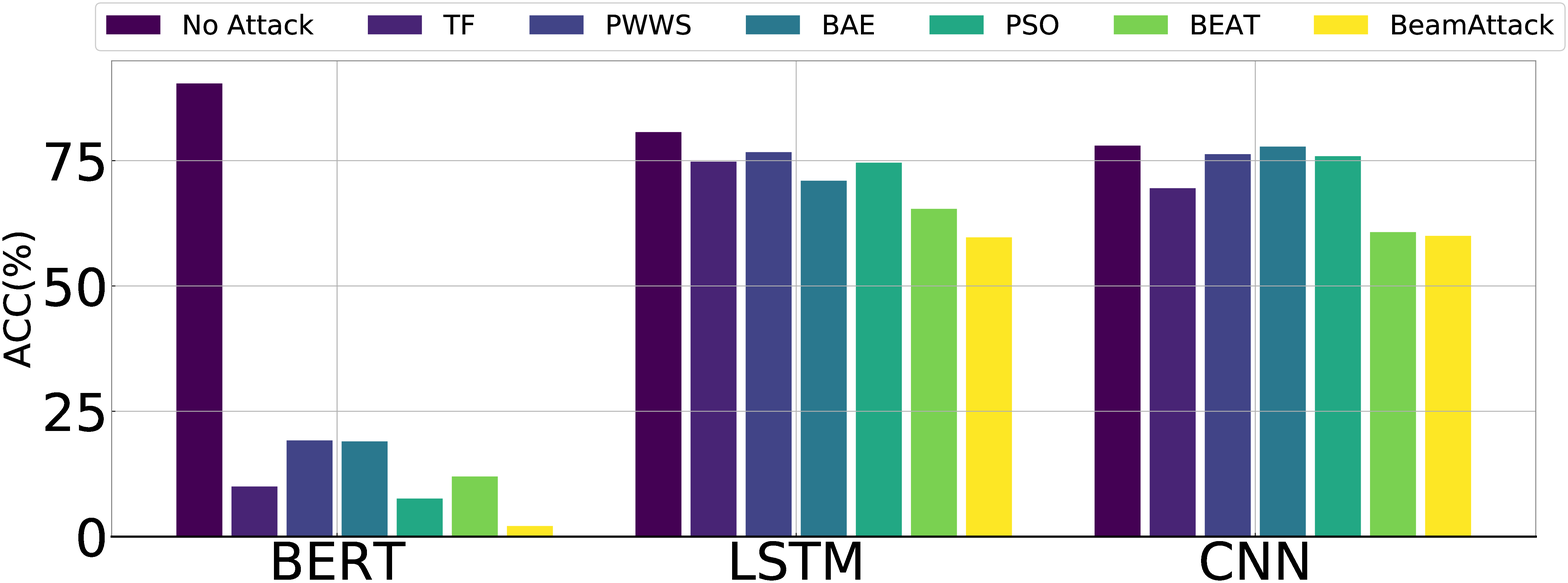}
  \vspace{-0.1in}
 \caption{Transfer attack on MR dataset. Lower accuracy indicates higher transferability (the lower the better).}
 \label{fig.trans}
 \end{figure}

 The transferability of adversarial examples reflects property that adversarial examples crafted by classifier $\mathcal{F}$ can also fool other unknown classifier $\mathcal{F'}$.  We evaluate the transferability on MR dataset across CNN,LSTM and BERT. In detail, we use the adversarial examples crafted for attacking BERT on MR dataset to evaluate the transferability for CNN and LSTM models. As shown in the Figure~\ref{fig.trans},  the adversarial examples generated by BeamAttack  achieve the higher transferability than baselines.
 
 \vspace{-0.05in}
 \subsection{Adversarial Training}
 Adversarial training is  a prevalent technique to improve the model's robustness by  adding some adversarial examples into train data. To validate this, we train the CNN model on the MR dataset and obtains 80.4\% test accuracy. Then we randomly generate  1000 MR adversarial examples to its training data and retrain the CNN model. The result is shown in the Table~\ref{tab.advtrain}, CNN model obtains 83.3\% test accuracy, higher than origin test accuracy. Although there is no significant change in ASR,  BeamAttack needs to replace more words and more model queries to attack successfully with WSR and model queries increasing.  It indicates that adversarial training effectively improves the generalization and robustness of the model.  
 
 \begin{table}[htbp]
 \centering
 \small
 %\scalebox{1}
 \caption{The performance of CNN with(out) adversarial training on the MR dataset.}
  \vspace{-0.1in}
 \label{tab.advtrain}
 \begin{tabular}{p{50pt}<{\centering}p{80pt}<{\centering}p{40pt}<{\centering}p{40pt}<{\centering}p{40pt}<{\centering}p{40pt}<{\centering}}\toprule[1.5pt]
                     &Origin ACC(\%)& ASR(\%)   & WSR(\%)  & SIM(\%)  & Query  \\\hline
 \textbf{Original}   &80.4 & 99.87 & 8.20  & 91.08 & 563.1  \\
 \textbf{Adv.Training}&\textbf{83.3} & \textbf{99.75} & \textbf{8.67} & \textbf{90.82} & \textbf{606.4}\\\bottomrule[1.5pt]
 \end{tabular}
 \end{table}
 \vspace{-0.05in}
\section{Conclusion}
In this paper, we propose an efficient adversarial textual attack algorithm BeamAttack. The BeamAttack makes full use of word embedding space and BERT to generate substitutions  and fine-tune beam search to expand search spaces. Extensive experiments demonstrate BeamAttack balances the attack success rate, qualify of adversarial examples and time consumption. In addition, the adversarial examples crafted by BeamAttack  are contextual-aware and improve models' robustness during adversarial training.
\bibliographystyle{splncs04}
\bibliography{ref}

\end{document}